\documentclass[manuscript]{acmart}
\usepackage{float}
\usepackage{longtable}
\usepackage{booktabs}
\usepackage{placeins}
\usepackage{soul,color}

\usepackage{color,soul}
\usepackage{svg}
\AtBeginDocument{%
  }

\setcopyright{acmlicensed}
\copyrightyear{2018}
\acmYear{2018}
\acmDOI{XXXXXXX.XXXXXXX}

\acmJournal{TOG}
\acmVolume{37}
\acmNumber{4}
\acmArticle{111}
\acmMonth{8}

\setcopyright{none}
\renewcommand\footnotetextcopyrightpermission[1]{}

\begin{document}

\title{Evaluating Federated Learning for Cross-Country Mood Inference from Smartphone Sensing Data}




\author{Sharmad Kalpande}
\orcid{0009-0003-7904-4881}
\affiliation{%
  \institution{Indian Institute of Science Education and Research Bhopal}
  \city{Bhopal}
  \country{India}
}
\email{kalpande22@iiserb.ac.in}

\author{Saurabh Shirke}
\orcid{0009-0000-7913-1959}
\affiliation{%
  \institution{Indian Institute of Science Education and Research Bhopal}
  \city{Bhopal}
  \country{India}
}
\email{shirke22@iiserb.ac.in}

\author{Haroon R. Lone}
\orcid{0000-0002-1245-2974}
\affiliation{%
  \institution{Indian Institute of Science Education and Research Bhopal}
  \city{Bhopal}
  \country{India}
}
\email{haroon@iiserb.ac.in}

\begin{abstract}
Mood instability is a key behavioral indicator of mental health, yet traditional assessments rely on infrequent and retrospective reports that fail to capture its continuous nature. Smartphone-based mobile sensing enables passive, in-the-wild mood inference from everyday behaviors; however, deploying such systems at scale remains challenging due to privacy constraints, uneven sensing availability, and substantial variability in behavioral patterns.

In this work, we study mood inference using smartphone sensing data in a cross-country federated learning setting, where each country participates as an independent client while retaining local data. We introduce FedFAP, a feature-aware personalized federated framework designed to accommodate heterogeneous sensing modalities across regions. Evaluations across geographically and culturally diverse populations show that FedFAP achieves an AUROC of 0.744, outperforming both centralized approaches and existing personalized federated baselines. Beyond inference, our results offer design insights for mood-aware systems, demonstrating how population-aware personalization and privacy-preserving learning can enable scalable and  mood-aware mobile sensing technologies
\end{abstract}

\maketitle

\section{Introduction}


Mood instability serves as a primary behavioral biomarker for several psychological conditions, including depression and bipolar disorder, as these disorders are characterized by frequent and sometimes abrupt changes in affective state~\cite{meegahapola2023generalization,saha2017inferring}. Since mood evolves continuously over time rather than remaining static, capturing these temporal fluctuations is critical for early detection and timely intervention. In this context, smartphone sensing often referred to as digital phenotyping, has emerged as a powerful paradigm within ubiquitous computing by enabling continuous, real-world observation of behavior~\cite{morshed2019prediction,busso2025diversityone}. Modern smartphones passively and persistently collect signals related to daily routines, mobility, physical activity, and device interaction, allowing mood-related changes to be monitored without interrupting users’ normal activities. By leveraging passive signals such as application usage patterns, geolocation variability, and physical activity levels, mobile sensing supports continuous mood inference without requiring active user input~\cite{gao2020n}. This substantially reduces user burden compared to manual self-reporting and provides a scalable foundation for delivering just-in-time and personalized mental well-being support.

Several prior studies have investigated mood inference using behavioral data, demonstrating that mood states can be inferred from observable patterns in daily activity and interaction~\cite{moodscope,cho2019mood}. In parallel, with the widespread adoption of mobile devices, several studies have leveraged smartphone sensing to model mood using passive signals such as mobility, physical activity, and phone usage~\cite{wang2014studentlife,xu2022globem,gao2020n,busso2025diversityone}. While these approaches report encouraging results, most prior work is conducted on data collected from a single geographic region or a specific population (e.g., a university cohort) and relies on centralized training settings in which all data are aggregated on a server~\cite{morshed2019prediction,zhang2019inferring}. A limited number of studies have explored cross-country or multi-region mood inference; however, these models are still trained centrally~\cite{meegahapola2023generalization,bangamuarachchi2025inferring}. Such centralized approaches often struggle to generalize across culturally diverse populations, as behavioral signals associated with mood can vary across regions, and they also raise privacy concerns due to the need for sharing sensitive personal data.

Centralized training is particularly problematic for smartphone-based mood inference because passive sensing data captures fine-grained information about individuals’ daily behavior, mobility, and interactions~\cite{shen2025passive,harari2016using}. Aggregating such data increases the risk of privacy breaches and misuse, concerns that are further amplified in cross-country settings where data-sharing regulations, institutional policies, and ethical constraints restrict the transfer of personal data across national boundaries~\cite{arefin2025securing}. Federated learning (FL) offers a privacy-preserving alternative by enabling collaborative model training without requiring raw smartphone data to leave local sites~\cite{li2020federated}. By keeping data stored locally within each country and sharing only model updates, FL aligns with real-world data governance constraints while still enabling learning from geographically and culturally diverse smartphone sensing data.

In this work, we investigate mood inference using smartphone sensing data in a federated learning setting~\cite{li2020federated}, where each country participates as an independent client in collaborative model training. We demonstrate that the proposed FedFAP federated architecture can be trained effectively in this setting, achieving an AUROC of 0.744, and outperforming existing federated baselines as well as centralized training setups evaluated in our study. By framing mood inference as a cross-country federated problem, we explicitly account for data imbalance, feature heterogeneity, and privacy constraints that arise in real-world deployments. Our study systematically evaluates the feasibility and effectiveness of federated learning for mood inference across culturally and geographically diverse populations. The key contributions of this work are summarized as follows:

\begin{itemize}
    \item \textbf{FedFAP (Feature-Aware Personalization)}, a personalized federated learning framework that enables clients to retain and leverage heterogeneous sensor modalities while supporting collaborative mood inference across distributed populations.
    \item A comprehensive cross-country evaluation of federated mood inference using smartphone sensing data, demonstrating that population-aware personalization can outperform both centralized learning and existing personalized federated baselines under feature heterogeneity and privacy constraints.
\end{itemize}

\section{Related Work}
\subsection{Mood Inference}
Affective state (i.e., mood and emotion) inference has been widely studied using controlled sensing modalities, particularly physiological and neural signals. Early studies demonstrated that affective states can be inferred from physiological measurements such as electrodermal activity, heart rate, respiration, and muscle activity using feature-based machine learning models trained on self-reported emotion labels \cite{picard2001toward,kim2008emotion}. With advances in neural sensing, electroencephalography (EEG) has been extensively used to model emotional states, where handcrafted or learned representations from multichannel EEG signals are employed to infer valence, arousal, or discrete emotion categories elicited through audiovisual stimuli \cite{koelstra2011deap,zheng2015investigating,chen2019accurate}. Subsequent work explored multimodal affect modeling by combining physiological signals with audio, motion, and linguistic cues, enabling richer representations of mood in more naturalistic settings~\cite{alhanai2017predicting}. Recognizing substantial inter-individual variability in affective responses, several studies emphasized personalized modeling strategies, particularly for mental health monitoring and depression assessment \cite{chatterjee2023towards,choi2022depressed}. More recent approaches have shifted toward continuous and longitudinal mood modeling using wearable and smartwatch-based sensing, integrating passive signals such as activity, sleep, heart rate, and daily rhythm patterns with self-reports to forecast affect over extended periods, including the use of movement and gait dynamics as non-intrusive indicators of emotional state \cite{jaques2017predicting,cho2019mood,yang2024integrating,quiroz2018emotion}.

\paragraph{Mobile sensing:} Building on this shift toward continuous and longitudinal affect modeling, smartphone sensing has further gained attention as a scalable means to passively capture daily behavioral patterns in real-world settings. Prior work has leveraged a wide range of smartphone-derived signals, including location, mobility, phone usage, communication logs, ambient context, and interaction patterns, combined with ecological momentary assessments or daily self-reports to infer mood and depressive states at varying temporal resolutions~\cite{pillai2024investigating}, ranging from hourly to daily inferences \cite{jacobson2020passive,ma2012daily,likamwa2013moodscope,asselbergs2016}. Several studies demonstrated that specific behavioral cues, such as mobility regularity derived from GPS traces or longitudinal phone usage patterns, can serve as effective proxies for depressive symptom changes, often employing personalized models to account for individual differences \cite{canzian2015trajectories,likamwa2013moodscope}. Large-scale longitudinal datasets such as StudentLife further highlighted the potential of continuous smartphone sensing for analyzing mood, stress, and mental well-being over extended periods in naturalistic academic settings \cite{wang2014studentlife}. More recent approaches explored richer sensing modalities, including typing dynamics and front-facing camera images, as well as deep learning–based multimodal fusion techniques, to improve mood and depression inference performance \cite{cao2017deepmood,nepal2024moodcapture}. Extensions incorporating wearable data alongside smartphone sensing have also been proposed to model mood instability and long-term mood trajectories using passive digital phenotypes related to activity, sleep, and daily rhythm patterns \cite{morshed2019prediction,cho2019mood}.

Despite promising results, many smartphone-based mood inference studies are conducted within limited populations or single geographic regions and rely on relatively small or homogeneous cohorts, which poses challenges for model generalization across users and contexts \cite{leaning2024smartphone,cho2019mood}. Recent cross-country studies have begun to explicitly examine these generalization issues, with large-scale datasets such as DiversityOne enabling mood inference across heterogeneous cultural settings and empirical analyses showing that country-specific or partially personalized models often outperform global models due to strong geographical and behavioral heterogeneity \cite{busso2025diversityone,meegahapola2023generalization,sahu2025anxiety}.

\subsection{Federated Learning (FL)}
FL is a collaborative machine learning approach where multiple clients train a shared model without sharing their raw data. Instead of centralizing data, each participant keeps data stored locally, performs training on their own device, and shares only the model updates (such as weights or gradients) \cite{li2020federated, mcmahan2017communication,kairouz2021advances,deng2025cross}. This architecture enables learning from distributed data while preserving privacy \cite{yang2019federated}. The effectiveness of FL depends critically on how client updates are aggregated at the server \cite{kairouz2021advances}. Widely adopted aggregation methods include Federated Averaging (FedAvg) \cite{mcmahan2017communication}, FedProx \cite{li2020federated_a}, and adaptive optimization approaches such as FedAdam \cite{reddi2020adaptive}, each designed to address challenges related to communication efficiency, convergence stability, and data heterogeneity across clients. However, these methods assume a single global model shared uniformly across clients, an assumption that often breaks down when data distributions differ substantially \cite{arivazhagan2019federated}. Personalized Federated Learning (PFL) addresses this limitation by enabling client adaptive models while preserving collaborative benefits \cite{tan2022towards}, with representative approaches including FedPer \cite{arivazhagan2019federated}, which shares representation layers while keeping task-specific layers local, and pFedMe \cite{t2020personalized}, which regularizes personalized client models toward a global reference.

This privacy by design approach is particularly suitable for mood inference and mental health applications, where behavioral signals and emotional states are highly sensitive \cite{rieke2020future}. Consequently, FL has been increasingly explored for mental health assessment across various conditions. Recent studies have applied FL to depression detection using mobile keystroke dynamics \cite{xu2021privacy}, speech analysis \cite{bn2022privacy,cui2022privacy}, and clinical electronic health records \cite{lee2023privacy}. Other work has addressed suicidal ideation detection from social media text \cite{ji2019detecting}, obsessive-compulsive disorder from wearable sensors \cite{kirsten2021sensor}, and bipolar disorder transition inference \cite{lee2023privacy}.

However, existing studies largely focus on discrete mental health conditions rather than modeling mood as a continuous and dynamic state. Broadly, prior approaches can be categorized into two main directions based on the type of data employed. The first direction explores physiological data–driven federated learning, leveraging biosignals such as ECG, EDA, EEG, and other physiological measurements collected from wearable or laboratory-grade sensors~\cite{nandi2022federated,gahlan2024federated}. For example, FedMultiEmo \cite{gul2025fedmultiemo} employs a federated learning framework for emotion recognition using physiological signals, demonstrating that decentralized training over biosensor data can effectively model affective states while preserving privacy. Similarly, recent studies have applied federated learning to emotion recognition using physiological biosignals such as EEG and EDA, showing the feasibility of privacy-preserving affect modeling without centralizing sensitive physiological data \cite{simic2024enhancing}. The second and more prevalent direction focuses on mobile-based data, including mobile sensing signals and interaction-based features such as keystroke dynamics. For instance, FedMood \cite{xu2021fedmood} employs a multi-view FL framework with late fusion strategies to handle temporal inconsistencies across modalities for depression score inference from BiAffect keyboard data. A recent work~\cite{martis2025federated} applied FL with continual learning to mood and depression inference, evaluating performance under realistic conditions such as client dropout, non-IID data distributions, and temporal drift. Similarly, other studies have explored privacy-preserving frameworks using on-device data such as speech signals and linguistic patterns for real-time mood assessment \cite{shin2023fedtherapist}. 

Despite these advances, existing federated mood inference systems share a critical limitation: they assume all clients possess identical sensor modalities \cite{khalil2024exploring}. The systematic review~\cite{khalil2024exploring} explicitly identified ``feature heterogeneity'' where clients have different feature sets due to varying sensor availability as a recognized challenge in FL for mental health, yet none of the 16 reviewed studies proposed architectural solutions to address this. Similarly, \cite{martis2025federated} achieved performance gap of about 10–15\% relative to centralized models using standard FedAvg aggregation, assuming homogeneous feature spaces across all clients.

Overall, existing FL studies in affective computing predominantly focus on mental health conditions such as depression or stress, while relatively few works explicitly address mood inference as a distinct and longitudinal task. Moreover, most prior approaches emphasize privacy preservation and feasibility, often relying on global models, homogeneous feature spaces. These limitations restrict applicability to scenarios where sensor availability varies due to device heterogeneity, user permissions, or regional technology adoption patterns. This gap is particularly problematic for cross-country deployments, where behavioral patterns, cultural norms, and sensor availability differ substantially across regions.

\section{Methodology}

Smartphone sensing data collected across different countries and cultural contexts naturally exhibits substantial heterogeneity~\cite{busso2025diversityone}. Behavioral patterns such as mobility, physical activity, social interaction, and smartphone usage are strongly influenced by cultural norms, environmental conditions, and daily routines. For example, step count patterns may vary across regions due to differences in transportation habits, urban infrastructure, and climate. Similarly, Bluetooth-based proximity signals can reflect distinct social interaction styles, while app usage and notification patterns are shaped by local technology practices~\cite{meegahapola2023generalization}. In addition, cultural factors may influence how individuals self-report affective states, meaning that similar sensed behaviors may correspond to different mood labels across regions. These variations introduce pronounced \textbf{non-IID characteristics} in both sensor data distributions and labels, making cross-country mood inference particularly challenging.

\subsection{Dataset} 

To study mood inference under such heterogeneous conditions, we use the DiversityOne dataset~\cite{busso2025diversityone}, a large-scale multi-country mobile sensing dataset designed to capture everyday behaviors through a combination of passive smartphone sensing and in-situ self-reports. DiversityOne contains 329,974 mood self-reports collected from 782 college students over a four week (28 day) longitudinal study, along with data from 26 smartphone sensor modalities. Data were collected across eight countries spanning multiple geographic and cultural regions, including China and Mongolia (Asia), India (South Asia), Mexico and Paraguay (Latin America), and Denmark, Italy, and the United Kingdom (Europe), representing diverse population profiles. In the study protocol, participants received mood prompts at regular intervals of approximately 30 minutes throughout the day. For our experiments, we used data from six countries: Denmark, China, India, Paraguay, United Kingdom, and Mexico. Data from the remaining two countries were excluded due to missing or corrupted sensor streams. The selected six countries provide sufficient data volume and participant coverage to support cross-country federated analysis while preserving the dataset’s geographic and cultural diversity.

Participants used the iLog mobile sensing application to provide repeated (every 30 minutes) in-the-moment self-reports, including mood ratings, while the application simultaneously recorded continuous sensing and interaction-based smartphone signals. For our work, we focus specifically on the mood labels, which were self-reported on a 5-point ordinal scale ranging from 1 (very positive) to 5 (very negative), with 3 representing a neutral mood, distribution of these labels is shown in Figure\ref{fig:5label}. The dataset provides a rich set of mobile sensing signals spanning both continuous sensing streams and interaction based modalities. Continuous sensing includes mobility, physical activity, and environmental context, such as motion derived features, activity recognition, step counts, location and mobility patterns, and connectivity signals (WiFi and Bluetooth). Interaction sensing captures immediate phone behaviors, including app usage, notifications, screen events, and touch activity. Together, these modalities provide a comprehensive view of daily behavior that can be linked to mental and emotional states.

\subsection{Pre-processing}

\subsubsection{Aligning self-reports and Passive Sensor Data}
Each mood self-report in the dataset is associated with a user identifier and a reported timestamp $T$. To align these labels with passive sensing data, we constructed a fixed-length sensor context window centered around $T$. Specifically, for each mood report we defined a 10 minute window spanning from $T - 5$ minutes to $T + 5$ minutes, i.e., a ``$T \pm 5$ minutes'' interval. This window acts as the reference period for extracting sensor measurements across all modalities. The choice of a 10 minute context window follows prior work on the DiversityOne protocol and related smartphone mood inference literature, where this duration was found to provide a good empirical trade-off between capturing sufficient behavioral context and avoiding excessive temporal drift between sensing patterns and the self-reported mood label~\cite{meegahapola2023generalization}.


\subsubsection{Label Mapping}
\begin{figure}[t]
    \centering
    \begin{minipage}[t]{0.48\linewidth}
        \centering
        \includegraphics[width=\linewidth]{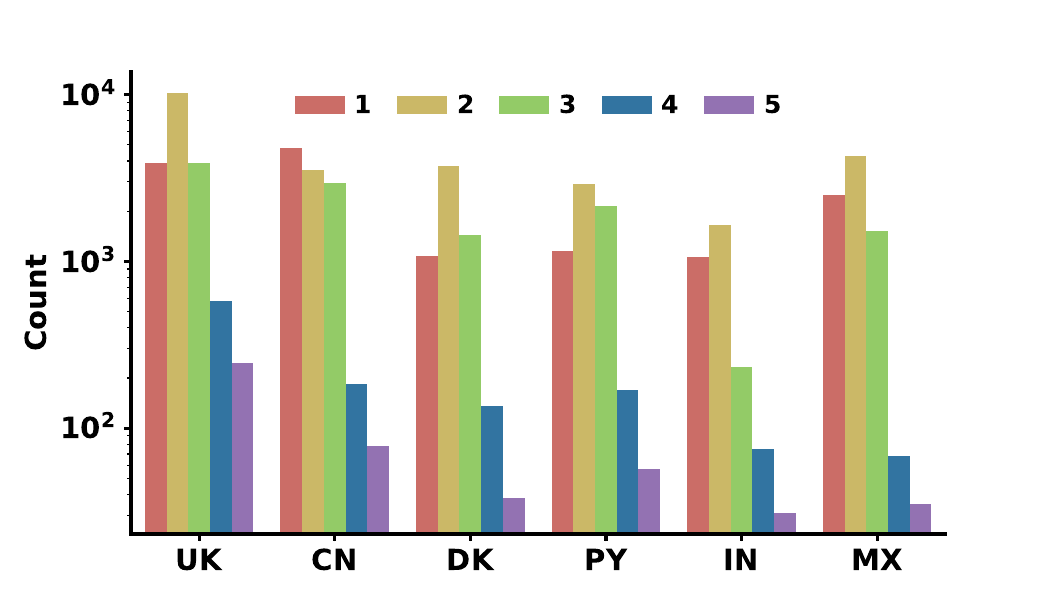}
        \caption{Distribution of samples across five label classes.}
        \label{fig:5label}
    \end{minipage}
    \hfill
    \begin{minipage}[t]{0.48\linewidth}
        \centering
        \includegraphics[width=\linewidth]{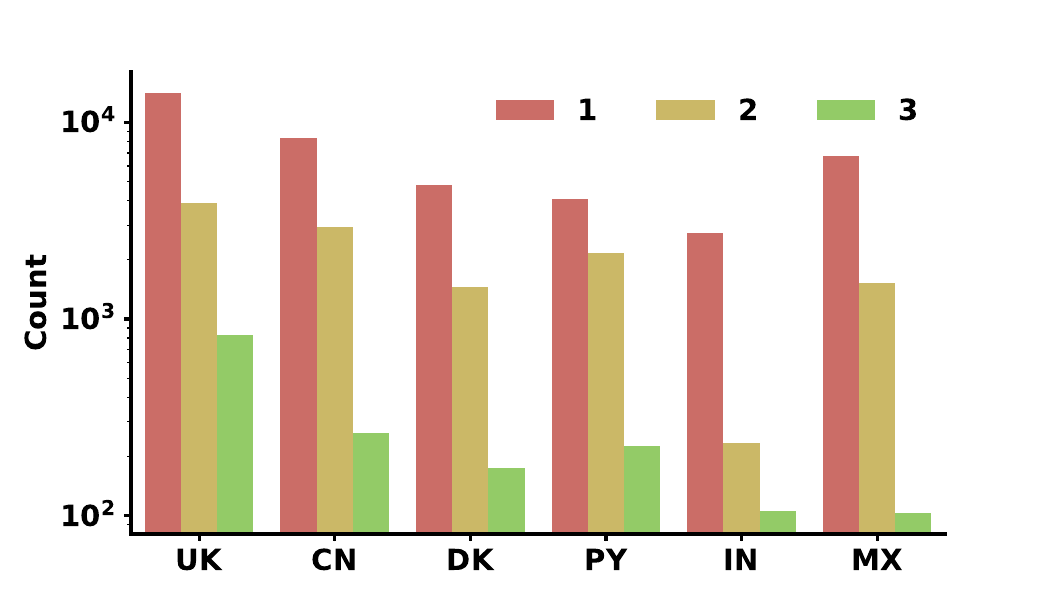}
        \caption{Distribution of samples across three label classes.}
        \label{fig:3label}
    \end{minipage}
\end{figure}

Mood self-reports in DiversityOne are originally provided on a 5-point ordinal scale ranging from 1 (very positive) to 5 (very negative). Inspection of the country-wise label distribution (see Figure~\ref{fig:5label}) reveals a clear imbalance across the five levels. In particular, labels corresponding to negative moods (4 and 5) are sparsely represented across all countries. For example, in Denmark (DK) and China (CN), label 4 contains approximately 100–200 samples, while label 5 contains fewer than 100 samples, compared to several 1000 samples for labels 1 and 2. A similar pattern is observed in India (IN) and Mexico (MX), where label 5 appears only a few dozen times and label 4 generally remains below a few hundred samples, whereas positive labels (1 and 2) dominate the distribution. This long-tailed pattern is consistent across countries, indicating that fine-grained five-class modeling would be highly sensitive to class imbalance and client-level sparsity, particularly for negative mood categories.

To address this issue, we reformulate the task into a three-class classification problem by grouping scores 1–2 as positive, 3 as neutral, and 4–5 as negative, which are subsequently mapped to labels 1, 2, and 3, respectively. The resulting class distribution after relabeling is illustrated in Figure~\ref{fig:3label}, providing a more stable and comparable learning setup across countries and clients while preserving the ordinal structure of the original mood reports. Table~\ref{tab:country_participants} summarizes the number of participants and the total number of mood self-reports for each country included in our analysis after preprocessing.

\begin{table}[h]
    \centering
    \caption{Participants and total instances (mood self-reports) across countries Participants and total instances (number of mood self-reports) across countries after preprocessing.}
    \begin{tabular}{lcc}
        \toprule
        \textbf{Country} & \textbf{Participants} & \textbf{Total Instances} \\
        \midrule
        CN    & 38 & 11523 \\
        DK  & 17 & 6424  \\
        IN    & 19 & 3071  \\
        MX   & 19 & 8400  \\
        PY & 22 & 6467  \\
        UK       & 52 & 18832 \\
        \bottomrule
    \end{tabular}
    \label{tab:country_participants}
\end{table}

\subsection{Feature Extraction}
For each sensor modality, a fixed set of handcrafted features was computed whenever at least one measurement was available within the corresponding 10 minute sensor window of a mood report. The feature design followed a summary statistics approach, capturing signal magnitude, variability, and categorical distributions depending on the modality. Table \ref{tab:features_explanation} summarizes the source files used and the corresponding features extracted for each modality.

\begin{table*}[!h]
\centering
\caption{Summary of all sensor modalities and extracted features. Here, min, max, mean, std, and var denote the minimum, maximum, mean, standard deviation, and variance, respectively. RSSI refers to the Received Signal Strength Indicator.Entropy captures state diversity within a time window.}
\begin{minipage}{\textwidth}
\centering
\begin{tabular}{l c p{9.8cm}}
\hline
\textbf{Sensor File} & \textbf{No. of Features} & \textbf{Extracted Features} \\
\hline

activities & 6 &
Fraction of time spent in: On Foot / In Vehicle / On Bicycle / Still / Tilting / Unknown activity   \\

bluetooth & 7 &
Number of unique Bluetooth device addresses; mean/min/max/std/var of RSSI; entropy of device addresses.\\

cellularnetwork & 6 &
Number of unique cell towers; mean/min/max/std of signal strength; Entropy of cell-tower usage.\\

location & 17 &
Centermost latitude and longitude (representing location); mean/min/max of latitude/longitude/altitude (capturing spatial coverage and elevation changes); mean/min/max/std of speed; Radius of gyration; Total distance traveled. \\

notification & 2 &
Number of posted notifications; Number of removed notifications. \\

proximity & 4 &
mean/std/min/max of proximity values. \\

screen & 7 &
Duration spent in each screen state; number of screen ON–OFF episodes (i.e., continuous screen-on usage periods followed by a screen-off event); mean/min/max/std of screen-on episode duration. \\

stepcounter & 1 &
Total steps within the time window, derived from cumulative counter increments. \\

stepdetector & 1 &
Count of step-trigger events within the time window.\\

touch & 1 &
Number of touch events. \\

userpresence & 2 &
Duration spent in each user-presence state. \\

Wi-Fi networks & 5 &
Number of unique WiFi device addresses; mean/min/max/std of RSSI. \\

\hline
\textbf{Total Features} & \textbf{59} & \\
\hline
\end{tabular}

\label{tab:features_explanation}
\end{minipage}
\end{table*}

After feature extraction, the modality-specific feature files were combined into a single multimodal dataset by aligning them on the keys \((userid,\,interval)\). Here, each \textit{interval} corresponds to a fixed 10-minute sensing window, represented using its associated start and end timestamps \((start\_time,\,end\_time)\). Since all modality wise feature files follow the same interval structure, we merged them using an outer join to preserve all available instances across sensors. Consequently, if a user had data for a given interval in a particular modality, the corresponding feature values were included in the merged dataset; otherwise, the entries for that modality remained missing (NaN). This strategy maintains consistent temporal alignment of multimodal features while preserving the overall sensing coverage across modalities.

\begin{figure}[]
    \centering
    \includegraphics[width=0.71\linewidth]{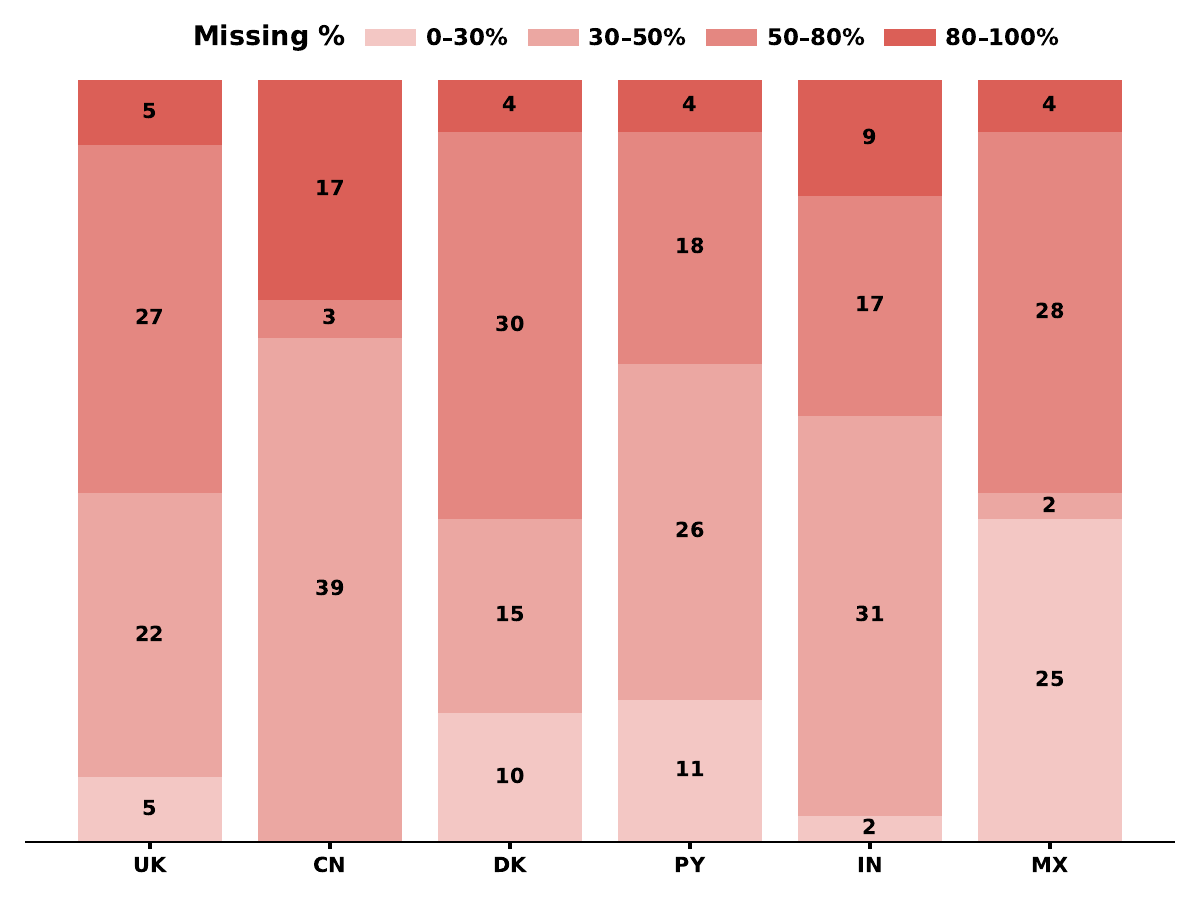}
    \caption{Distribution of feature missingness across countries, in which the marked areas denote the number of features belonging to different missingness ranges.}
    \label{fig:nan_handling}
\end{figure}

\subsubsection{Handling Nan Values}
\label{Handling_Nan}
After merging the modality-wise feature files into a single dataset, we observed that several feature columns contained missing values because not all sensors produced data for every user and every time interval. For example, a given 10 minute window may include Bluetooth proximity logs, while other modalities such as screen events or app usage may not record any events during the same interval, resulting in missing feature values. Figure \ref{fig:nan_handling} illustrates this pattern by showing, for each country, the proportion of features that fall into different missingness percentage ranges.  Based on this observation, we applied a two-stage strategy to handle missing data. First, features with excessive missingness were removed: any feature with 80\% or more NaN values (i.e., falling in the 80–100\% missingness range) within a country was discarded \cite{wei2018missing}. This step ensured that only features with sufficient coverage were retained and, due to differences in sensor availability across countries, resulted in slightly different feature sets for different countries.

Second, for the remaining features, missing values were imputed using KNN imputation~\cite{rashid2020predicting,meegahapola2023generalization} applied separately within each country, which estimates missing entries using similar samples from the same country-level dataset and helps preserve local structure. However, because features are first removed independently for each country based on missingness thresholds, the set of features retained for imputation differs across countries. As illustrated in Figure~\ref{fig:nan_handling}, countries exhibit distinct missingness profiles; for example, Denmark (DK) has 30 features falling in the 50–80\% missingness range, whereas Mexico (MX) retains 25 features in the low missingness range (0–10\%). Consequently, the final set of retained and imputed features differs across countries, resulting in country specific feature subsets and contributing to feature heterogeneity in the FL setup.


\subsubsection{Non-IID Data Characteristics}
The dataset exhibits strong non-IID characteristics across clients due to the following factors:
\begin{itemize}
    \item \textbf{Data and class imbalance:} The amount of data contributed by each country varies substantially, resulting in pronounced data imbalance across clients. As summarized in Table~\ref{tab:country_participants}, the UK and China (CN) contribute the largest numbers of mood self-reports, with 18,832 and 11,523 instances, respectively, while Mexico (MX) contributes 8,400 instances. In contrast, countries such as India (IN) contribute 3,071 instances, whereas Denmark (DK) and Paraguay (PY) contribute 6,424 and 6,467 instances, respectively. In addition to differences in data volume, the distribution of mood classes within each country is also skewed, with certain mood labels occurring far less frequently than others, as illustrated in Figure~\ref{fig:3label}, indicating the presence of class imbalance across clients.
    
    \item \textbf{Feature heterogeneity:} As described in Section \ref{Handling_Nan}, features were removed independently for each country based on missingness thresholds. As a result, the final retained feature sets differ across countries, which can also be observed in Figure~\ref{fig:nan_handling}. This leads to heterogeneous feature spaces, where certain sensor-derived features are present in some countries but absent or sparsely populated in others, yielding country-specific feature subsets.

\end{itemize}

\subsection{Proposed Method: FedFAP}\label{architetcture}

We adopt a FL framework in which each country is treated as an independent client participating in collaborative model training. All sensor-derived features and mood labels remain stored locally within each country, and only model updates are exchanged with a central server for aggregation. This setup enables collaborative learning across geographically distributed clients while avoiding centralized data collection.

\begin{figure}[t]
    \centering
    \includegraphics[width=0.75\linewidth]{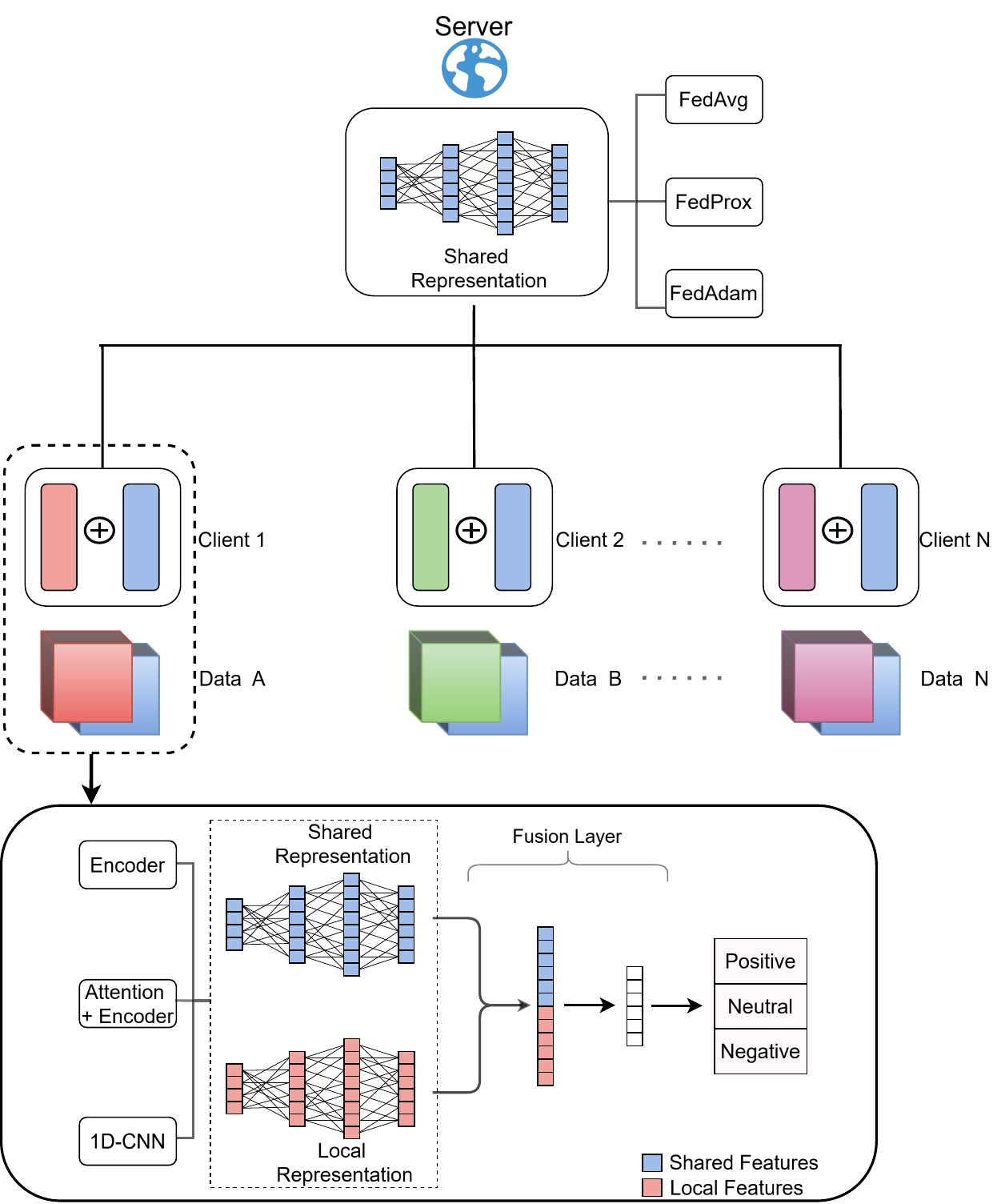}
    \caption{Proposed architecture comprising of a global server and N clients.}
    \label{fig:architecture}
\end{figure}

Our proposed framework, FedFAP, follows a personalized federated learning paradigm with an explicit separation between shared and client-specific representation learning (Figure~\ref{fig:architecture}). At the beginning of training, each client transmits a binary feature availability vector to the server, indicating which sensor modalities or features are consistently present in its local dataset. Using these vectors, the server computes the intersection of features that are reliably available across all participating clients and designates them as shared features. This shared feature set is communicated back to the clients and remains fixed throughout training

Each client maintains two neural representation encoders: a shared representation network, which operates exclusively on the server-identified shared features, and a local representation network, which processes the remaining client-specific features. The shared representation network is trained collaboratively across clients via federated aggregation at the server, while the local representation network remains private to each client. To preserve personalization and avoid cross network interference, gradients originating from the local representation network are not propagated back into the shared representation network.

The latent representations produced by the shared and local networks are subsequently combined through a lightweight fusion module, which integrates global and client specific information to generate task specific inferences. This separation between shared representation learning, local adaptation, and final fusion enables the model to capture globally consistent patterns while retaining sensitivity to client level heterogeneity, particularly in settings with multimodal data and uneven feature availability.

We instantiate this architecture using three different encoder designs—(i) simple feed forward encoders, (ii) attention based encoders, and (iii) one dimensional convolutional neural networks (1D CNNs), to evaluate the impact of representation capacity on personalization and generalization performance.

\begin{enumerate}
    \item \textbf{Feed-Forward Encoders:} In the feed forward configuration, both shared and local representation networks are implemented as multilayer perceptrons. The shared network comprises three fully connected layers with widths 128, 128, and 64, each followed by batch normalization, ReLU activation, and dropout (rate 0.3), producing a 64 dimensional shared latent representation learned via federated aggregation. The local network consists of two fully connected layers of width 128 with identical normalization, activation, and regularization, whose output is concatenated with the detached shared representation and projected to a 64 dimensional local embedding.
    
    The shared and local embeddings are concatenated and processed by a fusion module composed of two fully connected layers with widths 128 and 64, followed by a final classification layer. Batch normalization, ReLU activation, and dropout are applied after each hidden layer, and class probabilities are obtained via a softmax output.

\item {\bf Attention Based Encoders:} In the attention based configuration, the shared representation network extends the feedforward design by incorporating a residual feature attention mechanism. After two fully connected layers of width 128 with batch normalization and ReLU activation, a feature wise attention module applies a learnable, scaled residual transformation that reweights latent dimensions while preserving the original representation. This mechanism enables the model to emphasize informative features while suppressing less relevant ones, which is particularly beneficial under heterogeneous and partially observed feature distributions. The attended features are subsequently projected to a 64 dimensional shared embedding through a fully connected output layer with batch normalization and ReLU activation.

The local representation network follows the same feed forward structure as in the baseline configuration, producing a 64 dimensional local embedding after concatenation with the detached shared representation. To integrate shared and local information more selectively, the fusion module employs a gated attention mechanism that adaptively balances shared and local embeddings before classification. The fused representation is then passed through a multilayer classifier composed of fully connected layers with widths 128 and 64, followed by a final softmax layer for multiclass inference.

\item {\bf CNN Based Encoders:} In the CNN based configuration, both shared and local representation networks are implemented using one-dimensional convolutional layers to capture structured feature interactions, as illustrated in Figure \ref{fig:architecture}. The shared representation network consists of two 1D convolutional layers with kernel size 1 and channel widths 64 and 128, each followed by batch normalization and ReLU activation. The convolutional output is projected to a 64 dimensional shared embedding through a fully connected layer with dropout, batch normalization, and ReLU activation.

The local representation network follows a similar convolutional design, producing a latent feature representation that is concatenated with the detached shared embedding and mapped to a 64 dimensional local embedding via fully connected layers. The shared and local embeddings are then combined through a late fusion module composed of fully connected layers with widths 128 and 64, followed by a final softmax layer for multiclass inference. The convolutional encoders enable learning localized and structured feature transformations, which is beneficial when handling high dimensional or correlated feature sets.

\end{enumerate}

\subsubsection{\textbf{Federated Aggregation Setup}}\label{aggregation_methods}

In our framework, federated aggregation is applied exclusively to the parameters of the shared representation network at the server. During each communication round, clients train their local models using the current shared network parameters and return only the updated weights of the shared representation network to the server. The server aggregates these shared parameters to obtain a global shared model, which is then broadcast back to all clients for the next round of training. In contrast, the local representation network and the fusion module are optimized solely on the client and are never transmitted or aggregated. This design ensures that global knowledge is captured through the shared representation while client specific adaptations remain private and isolated.

\subsubsection{FedAvg}\label{FedAVG}
Federated Averaging (FedAvg) is a foundational aggregation method in FL, in which the server updates the global model by computing the arithmetic mean of the client updated shared model weights after each communication round \cite{mcmahan2017communication}. Each client trains the shared model locally and transmits only the updated shared parameters to the server for aggregation. Due to its simplicity and effectiveness, FedAvg is commonly used as a baseline for evaluating FL methods. The global model update in FedAvg is computed as:
\begin{equation}
    \mathbf{w}_{t+1} = \sum_{k=1}^{K} \frac{n_k}{n} \mathbf{w}_{t}^{k}
\end{equation}
Here, $\mathbf{w}_{t+1}$ denotes the aggregated global model at communication round $t+1$, obtained by averaging the locally updated models $\mathbf{w}_{t}^{k}$ from each participating client $k$, weighted by the proportion of data held by that client, $n_k$ is the number of samples in client $k$'s dataset, $n = \sum_{k=1}^{K} n_k$ is the total number of samples across all clients.

\subsubsection{FedProx}
FedProx follows the same server-side aggregation strategy as FedAvg, where the global model is updated by averaging client updates after each communication round. Its key distinction lies in the local optimization procedure, where a proximal regularization term is added to the client objective to limit excessive deviation from the global model, thereby improving robustness under data and system heterogeneity \cite{li2020federated_a}. At each communication round, clients receive the global model parameters from the server and use them as a fixed reference during local training. The local objective for client $k$ at round $t$ is defined as:

\begin{equation}
\min_{w_k} F_k(w_k) = f_k(w_k) + \frac{\mu}{2} \|w_k - w_t\|^2
\label{eq:fedprox}
\end{equation}
Here, $w_k$ denotes the local model parameters of client $k$, $w_t$ represents the global model parameters at round $t$, $f_k(w_k)$ is the local empirical loss on client $k$'s data, and $\mu \geq 0$ controls the strength of the proximal regularization. 

In our FedFAP framework, the proximal regularization is applied only to the shared representation parameters, while the local representation and fusion components remain unconstrained, enabling architectural personalization alongside improved training stability.

\subsubsection{FedAdam}
FedAdam extends federated aggregation by performing adaptive optimization at the server, rather than relying on simple parameter averaging \cite{reddi2020adaptive}. The server maintains first and second moment estimates of client updates across communication rounds. At round $t$, the server update proceeds as follows:

\begin{equation}
\begin{aligned}
\begin{aligned}
w_{t+1} &= w_t + \eta \cdot \frac{m_t}{\sqrt{v_t} + \epsilon}, \\
m_t &= \beta_1 m_{t-1} + (1 - \beta_1)\Delta_t
\end{aligned}
\qquad
\begin{aligned}
v_t &= \beta_2 v_{t-1} + (1 - \beta_2)\Delta_t^2, \\
\Delta_t &= \frac{1}{K} \sum_{k=1}^{K} (w_k^t - w_t)
\end{aligned}
\end{aligned}
\label{eq:fedadam}
\end{equation}

Here, the server updates the global model parameters $w_{t+1}$ using adaptive first and second moment estimates $m_t$ and $v_t$ computed from the aggregated client update $\Delta_t$. 
The parameters $\beta_1$ and $\beta_2$ control the decay rates of these moment estimates, $\eta$ is the server learning rate, and $\epsilon$ ensures numerical stability.

In our experiments, we use $\eta = 1 \times 10^{-3}$, $\beta_1 = 0.9$, $\beta_2 = 0.999$, and $\epsilon = 10^{-8}$. 
This adaptive aggregation improves convergence compared to standard averaging and is effective under non-IID data distributions.

\subsection{Baseline Personalized Federated Learning Methods}\label{baselines}
After feature extraction and dataset preparation, the resulting FL setting exhibits substantial client heterogeneity due to non-IID data distributions across clients. To provide a meaningful comparison and contextualize the effectiveness of the proposed framework, we evaluate our approach against established personalized federated learning baselines that are designed to handle client level data heterogeneity. These baselines represent widely adopted personalization strategies in FL and operate under the assumption of a shared feature space across clients.

\subsubsection{FedPer~\cite{arivazhagan2019federated}}
FedPer is a personalized federated learning baseline that partially shares model parameters across clients to mitigate data heterogeneity in non-IID settings. The model is decomposed into a shared feature encoder and a client specific classifier head, where only the encoder parameters are communicated and aggregated at the server, while the classifier head remains private to each client.

In our implementation, each client trains a neural network composed of a shared encoder and a private classification head. The encoder consists of three fully connected layers with 16, 32 and 16 hidden units respectively, each followed by Sigmoid activation, while the private head is a single layer with 16 input units and $C$ output units, where $C$ denotes the number of classes (three). Training is performed using the Adam optimizer with a learning rate of $1 \times 10^{-3}$ and cross-entropy loss. During each federated round, clients receive the current global encoder and return only the updated encoder parameters to the server for aggregation via FedAvg~\cite{mcmahan2017communication}, while the classifier heads are retained locally across rounds. As all clients share the same encoder architecture and input feature space, FedPer is applicable primarily in settings with a common feature set, where personalization is achieved through client specific classifier heads.

\subsubsection{pFedMe.~\cite{t2020personalized}}
pFedMe (personalized Federated Learning with Moreau Envelopes) is a personalized federated learning baseline that jointly learns a shared global model while enabling each client to obtain a personalized model adapted to its local data distribution. Personalization is achieved through a proximal coupling between the client-specific model and the global model, which constrains local solutions to remain close to the global parameters while allowing adaptation under non-IID data.

In our implementation, each client trains a fully connected neural network consisting of four layers: three hidden layers with 16, 32, and 16 units respectively (each followed by Sigmoid activation), and a final classification layer with $C$ output units, where $C$ denotes the number of classes (three). The entire network is personalized for each client through pFedMe's bi-level optimization. Optimization is performed using stochastic gradient descent with a personalized learning rate of $0.001$, a proximal coefficient $\lambda = 15.0$, and $K = 5$ inner-loop solver steps during local optimization.  At the start of each communication round, clients initialize from the received global model, perform personalized local optimization on each fold independently, and return the final personalized model after processing all folds. The server aggregates these personalized models using weighted averaging based on local data size to produce the updated global model. As all clients share the same model architecture and input feature space, pFedMe is applicable in settings with a common feature set, where personalization arises from client specific optimization rather than architectural separation.

\subsection{Evaluation and Training Configuration}\label{matrix}
Model performance was evaluated using a stratified leave-$n$-participants-out protocol implemented via 5-fold group-based cross-validation, where user identity was treated as the grouping variable~\cite{meegahapola2023generalization}. In each fold, approximately 20\% of participants were held out for testing, ensuring that evaluation was performed on entirely unseen users rather than additional samples from the same individuals. This setup reflects a realistic deployment scenario and prevents information leakage across training and testing splits.

Due to the presence of class imbalance in the multi-class mood labels (Figure~\ref{fig:3label}), macro-averaged AUROC was used as the primary evaluation metric. AUROC provides a threshold-independent measure of performance and treats all classes equally, making it appropriate for imbalanced multi-class settings. In addition, we report accuracy and weighted F1-score, where class contributions are weighted by their support, to better account for class imbalance while summarizing overall classification performance.

All FL experiments were conducted with a batch size of 64 and local learning rate of $1 \times 10^{-3}$ across all clients and communication rounds. To analyze the impact of local computation and communication frequency, we evaluated combinations of local training epochs ($E$) and federated rounds ($R$), including $(E,R) \in \{(5,5), (5,10), (10,10), (10,20), (10,50)\}$. Unless otherwise stated, the same evaluation protocol, training configuration, and hyperparameter settings were applied consistently across all baseline methods, aggregation strategies, and proposed model variants.

\subsection{Centralized Baselines and Evaluation Protocol}\label{centralized_matrix}
In addition to FL setup, we also evaluate centralized machine learning and deep learning baselines for comparison. In the centralized setting, data from all countries are assumed to be available on a single server and are used jointly for model training. To ensure a fair comparison with the federated setup, we follow a country-aware evaluation protocol. For a given target country, its data are first split into five stratified folds at the participant level. The model is trained using data from four folds of the target country together with data from all other countries, and evaluated on the remaining fold of the target country (same as Section~\ref{matrix}). This process is repeated across all five folds, enabling cross-validation for each country while maintaining consistent training and testing conditions across centralized and federated experiments.

For centralized mood inference, we evaluated a set of classical machine learning and deep learning models, using only the common feature set shared across all countries to ensure a consistent feature space. Logistic Regression (LR) was trained with a maximum of 1000 optimization iterations and balanced class weights to handle class imbalance. A Random Forest (RF) classifier with 300 decision trees was employed, using balanced class weighting and a fixed random seed for reproducibility. XGBoost (XGB) was trained using a multiclass soft probability objective with 300 boosting rounds, a learning rate of 0.05, and a maximum tree depth of 6, with row and column subsampling ratios set to 0.8. In addition, a one-dimensional convolutional neural network (1D-CNN) was trained for 50 epochs using the Adam optimizer with a learning rate of 0.001 and a batch size of 64, consisting of two convolutional layers with 32 and 64 filters followed by a fully connected layer and a softmax output, optimized using cross-entropy loss.

\section{Results}
\subsection{Comparison of FedFAP with Federated Baseline Methods}\label{Compare_baseline}
We first compare the proposed FedFAP framework against established personalized federated learning baselines, namely FedPer and pFedMe. All methods are evaluated under the same training configuration and evaluation protocol described in Section~\ref{baselines}, ensuring a fair and consistent comparison. For FedFAP, federated aggregation is performed using the standard FedAvg strategy, and a feed-forward encoder architecture is used for representation learning. Table~\ref{tab:baselines_comparisons} reports the comparative performance of all methods across the evaluated settings.

\begin{table}[]
\centering
\small
\begin{tabular}{lcccccccc}
\toprule
\textbf{Case} 
& \multicolumn{2}{c}{\textbf{5E 5R}} 
& \multicolumn{2}{c}{\textbf{5E 10R}} 
& \multicolumn{2}{c}{\textbf{10E 10R}} 
& \multicolumn{2}{c}{\textbf{10E 20R}} \\
\cmidrule(lr){2-3} \cmidrule(lr){4-5} \cmidrule(lr){6-7} \cmidrule(lr){8-9}
& \textbf{F1-w} & \textbf{AUROC}
& \textbf{F1-w} & \textbf{AUROC}
& \textbf{F1-w} & \textbf{AUROC}
& \textbf{F1-w} & \textbf{AUROC} \\
\midrule

FedPer 
& 0.663 & 0.512 
& 0.663 & 0.516 
& 0.667 & 0.520 
& 0.665 & 0.518 \\

\addlinespace[2pt]

pFedMe 
& 0.662 & 0.494 
& 0.662 & 0.512 
& 0.662 & 0.518 
& 0.662 & 0.491 \\

\addlinespace[4pt]
\cmidrule(lr){1-9}

\textbf{FedFAP} 
& \textbf{0.686} & \textbf{0.553} 
& \textbf{0.689} & \textbf{0.577} 
& \textbf{0.693} & \textbf{0.591} 
& \textbf{0.704} & \textbf{0.627} \\

\bottomrule
\end{tabular}
\caption{Comparison of baseline personalization methods with the proposed FedFAP under different training configurations (E: local epochs, R: communication rounds, F1-w: weighted F1 score).}
\label{tab:baselines_comparisons}
\end{table}

From Table~\ref{tab:baselines_comparisons}, it can be observed that FedFAP consistently outperforms the baseline federated methods (FedPer and pFedMe) across all evaluated settings. In particular, FedFAP achieves the highest AUROC scores in every configuration, with performance improving as the number of local epochs and communication rounds increases. For example, under the 10E–20R setting, FedFAP attains an AUROC of 0.627, compared to 0.518 and 0.491 for FedPer and pFedMe, respectively, while also achieving the best F1-weighted score (0.704). Similar trends are observed across other settings, where FedFAP consistently yields higher accuracy and F1-weighted scores. These results indicate that FedFAP more effectively captures personalized and cross-client representations under heterogeneous data distributions. Based on its consistent performance gains across all evaluated metrics and settings, we proceed with the proposed FedFAP method for all subsequent experiments.

\subsection{Evaluation of FedFAP with Different Encoder Architectures}\label{Compare_architecture}
We next evaluate the impact of different representation encoder architectures within the proposed FedFAP framework. Specifically, we compare the feed-forward, attention-based, and CNN based encoders described in Section~\ref{architetcture}. To ensure a fair comparison, all architectural variants are evaluated under the same training configuration and evaluation protocol, and FedAvg~\cite{mcmahan2017communication} is used as the aggregation method across all experiments. The comparative performance of the different architectures is reported in Table~\ref{tab:architeture_results}.




\begin{table}[h]
\centering
\small
\begin{tabular}{lcccccccc}
\toprule
\textbf{Case} 
& \multicolumn{2}{c}{\textbf{5E 10R}} 
& \multicolumn{2}{c}{\textbf{10E 10R}} 
& \multicolumn{2}{c}{\textbf{10E 20R}} \\
\cmidrule(lr){2-3} \cmidrule(lr){4-5} \cmidrule(lr){6-7}
& \textbf{F1-w} & \textbf{AUROC}
& \textbf{F1-w} & \textbf{AUROC}
& \textbf{F1-w} & \textbf{AUROC} \\
\midrule

Encoder
& 0.689 & 0.577
& 0.693 & 0.591
& 0.704 & 0.627 \\

\addlinespace[2pt]

Attn + Encoder
& 0.689 & 0.576
& 0.691 & 0.592
& 0.706 & 0.637 \\

\addlinespace[4pt]
\cmidrule(lr){1-7}

\textbf{1D-CNN}
& \textbf{0.696} & \textbf{0.606}
& \textbf{0.694} & \textbf{0.603}
& \textbf{0.725} & \textbf{0.684} \\

\bottomrule
\end{tabular}
\caption{Performance comparison of \textbf{Encoder}, \textbf{Attention+Encoder}, and \textbf{1D-CNN} under different training configurations (E: local epochs, R: communication rounds, F1-w: weighted F1 score).}
\label{tab:architeture_results}
\end{table}

From Table~\ref{tab:architeture_results}, we observe that the 1D-CNN–based architecture consistently achieves superior AUROC performance compared to both the Encoder-only and Attention+Encoder variants across all training settings. While the Encoder and Attention+Encoder configurations show comparable behavior, their AUROC gains remain relatively limited. In contrast, the CNN based design yields substantially higher AUROC values, improving from 0.606 under the 5E–10R setting to 0.684 under the 10E–20R setting, indicating stronger discriminative capability. This trend is consistent across all evaluated settings and is accompanied by improvements in F1-weighted scores, indicating more balanced class-wise performance. Based on these results, the 1D-CNN is selected as the final architecture for the proposed FedFAP method.

\subsection{Evaluation of Aggregation Methods in FedFAP}\label{Compare_aggregation}

We next analyze the impact of different federated aggregation strategies on the performance of the proposed FedFAP framework. Specifically, we compare FedAvg, FedProx, and FedAdam, as described in Section~\ref{aggregation_methods}. To ensure a controlled comparison, all experiments in this subsection use the CNN based encoder selected in the previous analysis and follow the same training configuration and evaluation protocol outlined in Section~\ref{matrix}. The comparative results for different aggregation methods are reported in Table~\ref{tab:avg_methods_results}.



 


\begin{table}[h]
\centering
\begin{tabular}{llcc}
\toprule
\textbf{Setting} & \textbf{Averaging Method} & \textbf{F1-weighted} & \textbf{AUROC} \\
\midrule
 & FedAvg  & 0.688 & 0.567 \\
5E 5R & FedProx & \textbf{0.689} & 0.563 \\
 & FedAdam & \textbf{0.689} & \textbf{0.569} \\

\midrule
 & FedAvg  & \textbf{0.696} & \textbf{0.606} \\
5E 10R & FedProx & 0.691 & 0.595 \\
 & FedAdam & 0.693 & 0.593 \\

\midrule
 & FedAvg  & 0.694 & 0.603 \\
10E 10R & FedProx & \textbf{0.705} & 0.623 \\
 & FedAdam & 0.699 & \textbf{0.624} \\
 
\midrule
 & FedAvg  & \textbf{0.725} & \textbf{0.684} \\
10E 20R & FedProx & 0.719 & 0.674 \\
 & FedAdam & \textbf{0.725} & 0.668 \\

\midrule
 & FedAvg  & \textbf{0.761} & \textbf{0.744} \\
10E 50R & FedProx & 0.755 & 0.732 \\
 & FedAdam & 0.748 & 0.724 \\
\bottomrule
\end{tabular}
\caption{Performance comparison of FedFAP across different federated aggregation methods.}
\label{tab:avg_methods_results}
\end{table}

From Table~\ref{tab:avg_methods_results}, we observe that the choice of federated aggregation method has a noticeable impact on performance, although no single method uniformly dominates across all training settings. In lower communication regimes (e.g., 5E–5R), all three methods exhibit comparable AUROC values, with FedAdam achieving a marginally higher AUROC (0.569) than FedAvg (0.567) and FedProx (0.563). As training progresses to 5E–10R, FedAvg attains the highest AUROC (0.606), outperforming FedProx (0.595) and FedAdam (0.593). A similar pattern is observed under 10E–10R, where FedAdam slightly outperforms FedAvg in AUROC (0.624 vs. 0.603), indicating that adaptive optimization can be beneficial under moderate communication budgets. However, under more intensive training settings, FedAvg consistently yields stronger discriminative performance. In particular, at 10E–20R, FedAvg achieves the highest AUROC (0.684) compared to FedProx (0.674) and FedAdam (0.668), and this gap further widens at 10E–50R, where FedAvg reaches an AUROC of 0.744, outperforming FedProx (0.732) and FedAdam (0.724). While F1-weighted scores follow similar trends, these results suggest that FedAvg provides more stable and scalable performance as communication rounds increase. Given the mixed behavior across settings and the pronounced gains at 10E–50R, we further analyze client-wise (country-level) performance under this configuration to better understand the source of performance differences across clients.

\begin{figure}[H]
    \centering
    \includegraphics[width=0.75\linewidth]{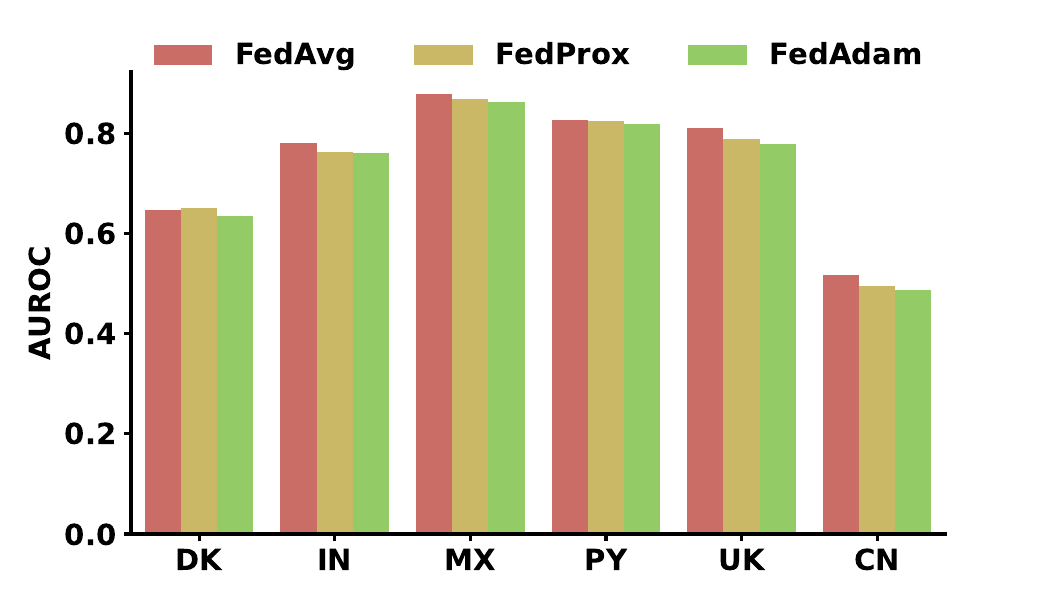}
    \caption{AUROC scores across different Federated averaging methods at 10 epochs and 50 rounds.}
    \label{fig:10E50R}
\end{figure}

Figure~\ref{fig:10E50R} presents the country wise AUROC comparison of different federated aggregation methods under the 10E–50R setting. Overall, FedAvg achieves strong and largely comparable AUROC performance across all countries, demonstrating stable behavior under cross-country data heterogeneity. While FedProx slightly outperforms FedAvg in the case of Denmark (DK), the performance of FedAvg remains comparable in this setting and higher across the remaining countries. Across all regions, FedProx consistently performs better than FedAdam, with FedAdam yielding lower AUROC values in comparison. Taken together, these results indicate that although FedProx can offer marginal gains for specific clients, FedAvg provides the most consistent and robust overall performance, supporting its use in subsequent experiments.

\subsection{Comparison of FedFAP with Centralized Baselines}
Having identified the optimal federated configuration for our evaluation, where the proposed FedFAP method consistently outperforms baseline approaches (Section~\ref{Compare_baseline}), the CNN based architecture achieves superior performance over encoder only and attention-based variants (Section~\ref{Compare_architecture}), and FedAvg emerges as the most effective aggregation strategy compared to FedProx and FedAdam (Section~\ref{Compare_aggregation}), we now turn to a comparison between centralized and FL setups. This comparison aims to assess whether the performance gains achieved through personalization and decentralized training are comparable to, or exceed, those obtained under centralized data sharing. Specifically, we evaluate centralized and federated models across countries, following the centralized evaluation protocol described in Section~\ref{centralized_matrix} and adopting the optimized FedFAP configuration with 10 local epochs and 50 communication rounds, ensuring a fair and consistent comparison across learning paradigms.
\begin{figure}[h]
    \centering
    \includegraphics[width=0.75\linewidth]{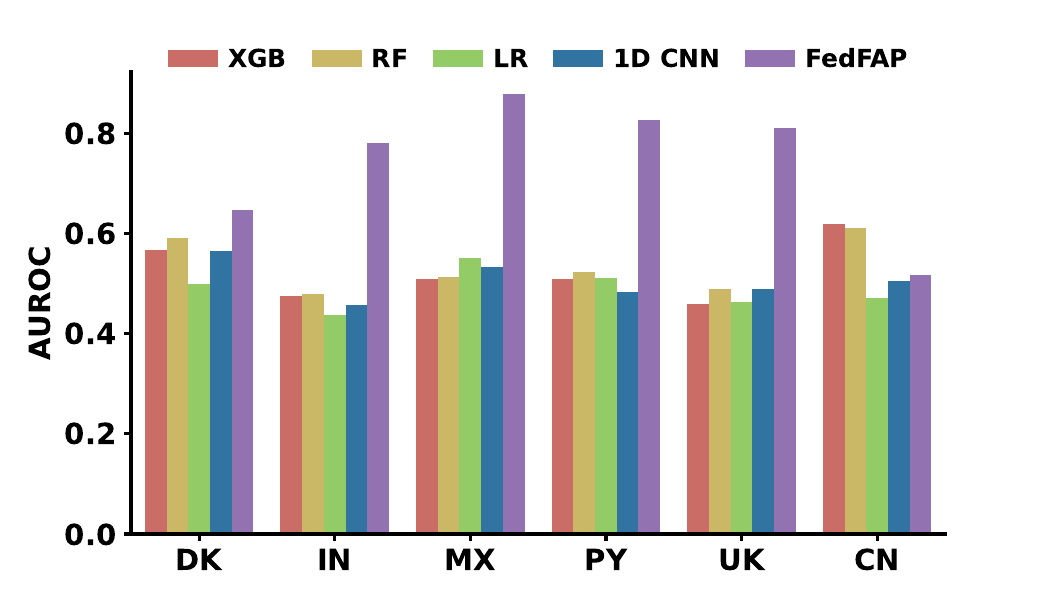}
    \caption{AUROC (y-axis) across countries (x-axis) for centralized baselines (XGBoost, RF, LR, 1D CNN) and the FedFAP.}
    \label{fig:FLvsCentralize_results}
\end{figure}
From Figure~\ref{fig:FLvsCentralize_results}, we observe that FedFAP achieves the highest AUROC across most countries, notably in India (IN), Mexico (MX), Paraguay (PY), and the UK, where it clearly outperforms all centralized baselines. In these regions, FedFAP attains AUROC values in the range of $\approx 0.78-0.88$, whereas centralized models typically remain below $\approx 0.55$, indicating a clear performance margin. In the case of Denmark (DK), FedFAP remains competitive and achieves an AUROC of approximately $\approx 0.65$, exceeding the best-performing centralized models, which achieve AUROC values of around $\approx 0.59-0.60$. An exception is observed for China (CN), where centralized approaches such as XGB and RF achieve higher AUROC values ($\approx 0.61-0.62$) compared to FedFAP ($\approx 0.52$). Despite this exception, FedFAP demonstrates consistently strong performance across countries and outperforms centralized learning methods in the majority of cases, highlighting its effectiveness in handling cross-country data heterogeneity without centralized data sharing.

\section{Discussion}

\subsection{Design Implications} 
Mood aware systems built on passive mobile sensing increasingly aim to operate in real world, longitudinal settings where users differ widely in behavior, context, and cultural norms. Our findings highlight that affective signals captured through smartphones are inherently heterogeneous and shaped by local routines, social practices, and usage patterns, making uniform modeling assumptions difficult to sustain at scale. Rather than focusing solely on improving global generalization under centralized learning, these results motivate a shift toward privacy-preserving, adaptive, and context sensitive system design. In the following, we discuss key design implications for mood-aware systems, focusing on (1) privacy-preserving learning for sensitive affective data, (2) robustness to feature heterogeneity and missing sensing modalities, and (3) population-aware personalization as an alternative to enforcing globally uniform behavioral models.

\subsubsection{Privacy Preserving Design for Mood-Aware Systems}
Passive smartphone sensing for mood inference relies on behavioral signals that are inherently personal and context rich, including location trajectories, phone usage routines, proximity patterns, and indicators of social presence. Unlike traditional approaches based on physiological data and self-report surveys \cite{shu2018review}, these signals can implicitly reveal sensitive aspects of an individual’s daily life, including mobility habits \cite{de2013unique}, social engagement \cite{eagle2006reality}, work routines, and lifestyle regularities. As such, the centralized collection and aggregation of behavioral sensing data raise significant privacy \cite{eagle2009inferring,payton2023privacy} and governance concerns \cite{voss2019cross}, particularly when data are shared across institutional or national boundaries. These concerns are further compounded by increasing regulatory diversity across countries, differences in cultural expectations around privacy \cite{payton2023privacy}, and growing emphasis on data sovereignty and local data governance \cite{voss2019cross}.

In cross-country mood sensing scenarios, centralized learning frameworks implicitly assume that sensitive behavioral data can be pooled and processed in a single location, an assumption that may not align with institutional policies, regulatory requirements, or user trust expectations. Behavioral data derived from smartphones are especially sensitive because even partial access, such as coarse location traces or communication metadata, can enable unintended inference of personal routines~\cite{de2013unique} or social relationships~\cite{mader2024learning}. Consequently, the risk associated with data breaches or misuse is not limited to individual privacy but may extend to broader ethical and societal implications~\cite{zuboff2023age}.

Considering these privacy and governance challenges, decentralized learning paradigms naturally come into focus. Our results suggest that privacy preserving system designs need not come at the cost of predictive performance. Despite operating without centralized data sharing, the proposed FedFAP framework achieves performance that is comparable to, and in many cases exceeds, centralized baselines across multiple countries. This indicates that decentralized learning can support cross-country mood inference while respecting local data governance constraints. From a design perspective, this shifts privacy from being a deployment limitation to an enabling constraint, encouraging mood-aware systems that are both privacy aligned and practically deployable across regions with differing regulatory and cultural expectations~\cite{bonawitz2019towards}.

\subsubsection{User and Region Dependent Feature Heterogeneity}
Feature heterogeneity poses a central challenge for mood inference, as the availability and reliability of sensing modalities vary across users and regions. Much of the existing work has focused on training and evaluating models within a single geographic region or institutional context~\cite{wang2014studentlife,leaning2024smartphone,cho2019mood}, where sensing configurations and user behavior are relatively consistent. Even in cross-country efforts, such as large scale mobile sensing datasets collected across multiple regions~\cite{busso2025diversityone}, analyses~\cite{meegahapola2023generalization} typically rely on a shared subset of common features to ensure a consistent input space across participants or countries. While this design choice simplifies modeling, it implicitly assumes uniform sensor availability and logging behavior, which may not hold in real-world deployments.

In practice, mobile sensing data are inherently incomplete and uneven. Smartphones may not log all sensing modalities continuously due to operating system constraints, battery optimization policies, user permissions, or device-specific limitations~\cite{stisen2015smart}. Moreover, social and cultural differences across countries can lead to systematic variation in the availability and reliability of certain modalities—for example, differences in location-sharing behavior, communication patterns, or app usage. Regulatory and institutional constraints may further restrict the collection or sharing of specific sensing signals across regions~\cite{busso2025diversityone}. As a result, enforcing a globally shared feature space often requires discarding locally available modalities, thereby removing potentially informative behavioral signals and limiting model expressiveness.

Our results suggest that such simplifications may come at a measurable performance cost. By design, the proposed FedFAP framework allows each client or country to retain and exploit its locally available feature modalities, without requiring alignment to a globally uniform feature space. Rather than dropping non-common features, FedFAP enables local representations to capture region-specific and user specific behavioral cues while still benefiting from shared learning across populations. The consistently higher AUROC achieved by FedFAP compared to both centralized and personalized federated baselines indicates that local feature richness plays a critical role in mood inference, particularly in heterogeneous, cross-country settings. From a system design perspective, these findings highlight the importance of embracing feature heterogeneity rather than suppressing it. Instead of enforcing strict feature uniformity across users or regions, mood aware systems can be designed to support adaptive and locally expressive representations that reflect real-world sensing constraints. Such an approach is better aligned with the realities of mobile sensing deployments, where missing modalities and uneven data availability are the norm rather than the exception.

\subsubsection{Rethinking Generalization in Mood-Aware Systems}
Most centralized and federated mood inference approaches aim to learn a single generalized model that can operate across diverse users, datasets, or geographic contexts. While this objective aligns with traditional machine learning practice, it may not be well suited for mobile sensing–based mood inference\cite{bangamuarachchi2025inferring,taylor2017personalized,sahu2025anxiety}, where behavioral signals are inherently personal and shaped by individual routines, habits, and social contexts. Smartphone-derived data reflect how a specific person interacts with their device and environment, and these patterns can vary substantially across individuals and populations. As a result, enforcing strong global generalization may overlook meaningful local structure in the data.

Several prior studies have attempted to address this challenge through domain adaptation, where models are trained on one dataset or population and evaluated on another~\cite{sano2015recognizing,meegahapola2023generalization,xu2022globem}. However, such approaches implicitly assume that behavioral representations learned from one group can reliably transfer to another, an assumption that may not hold in cross-country or cross-cultural settings where sensing behavior and data availability differ substantially~\cite{meegahapola2024m3bat,adler2022machine}. Our findings suggest that this challenge is better framed as a personalization problem rather than a pure generalization problem.

The performance of the proposed FedFAP framework supports this perspective. By enabling personalized local representations while still benefiting from shared learning across populations, FedFAP achieves consistently higher AUROC compared to both centralized models and existing personalized federated baselines. This indicates that mood inference benefits more from population-aware personalization than from enforcing globally uniform behavioral models. From a design standpoint, these results motivate a shift away from treating generalization as the primary goal and toward developing mood-aware systems that explicitly embrace personalization as a core requirement for robust, real-world deployment.

\subsection{Insights from FedFAP evaluation}
\subsubsection{Comparison with Personalized Federated Baselines}
In the landscape of cross-country mood inference using mobile sensing, the proposed FedFAP framework demonstrates consistently stronger performance compared to established personalized FL baselines such as FedPer and pFedMe \cite{arivazhagan2019federated,t2020personalized} across all evaluated configurations (Table \ref{tab:baselines_comparisons}). While these methods introduce personalization through classifier-level adaptation or client-specific optimization, their personalization mechanisms primarily operate at the parameter or objective level, without explicitly accounting for heterogeneity in the underlying feature space. In contrast, FedFAP enables clients to retain and exploit locally available feature modalities, allowing region- and user-specific behavioral signals to contribute directly to representation learning. This distinction becomes particularly important in cross-country settings, where sensing availability and data characteristics differ across clients, and helps explain the performance gains observed with FedFAP over existing personalized federated baselines.

\subsubsection{Why FedFAP Outperforms Centralized Models}

FedFAP achieves higher AUROC than centralized baselines in five out of six evaluated countries (Figure \ref{fig:FLvsCentralize_results}), demonstrating that personalized federated learning can outperform conventional centralized training in cross-country mood inference. This improvement arises from FedFAP’s architectural design, which combines a shared encoder for learning common representations across countries with a local encoder that captures client-specific patterns. This enables each country to benefit from collaborative learning while retaining personalization, which is critical under heterogeneous data distributions.

An exception is observed for China (CN), where centralized XGBoost and Random Forest outperform FedFAP (Figure \ref{fig:FLvsCentralize_results}). This behavior is primarily due to substantial feature loss during preprocessing: 17 features were removed from the Chinese dataset because they exceeded the 80\% missingness threshold (Section \ref{Handling_Nan}). Since FedFAP’s local encoder relies on client-specific features for personalization, the reduced feature set limits its ability to learn meaningful local representations. In contrast, centralized models are trained only on the common feature subset shared across all countries and therefore are not affected by this loss of client-specific information.

Overall, these results indicate that FedFAP’s advantage over centralized models stems from its ability to exploit locally available features for personalization while simultaneously learning shared representations across clients. However, when a client suffers from severe feature sparsity, the benefits of personalization diminish, and centralized models trained on homogeneous feature sets may become comparatively stronger.

\subsubsection{Impact of Encoder Architectures}
The 1D-CNN encoder achieves the highest performance among the three evaluated architectures, consistently outperforming both the feed-forward encoder and the attention-enhanced encoder across all training configurations (Table~\ref{tab:architeture_results}). Prior work in mobile and physiological sensing has demonstrated that convolutional architectures are effective at extracting structured patterns from sensor-derived data~\cite{zeng2014convolutional}. More broadly, 1D-CNNs are widely used for their ability to learn hierarchical representations from structured inputs~\cite{kiranyaz20211d}.

In our federated setting, this architectural choice proves particularly effective because 1D-CNNs can exploit local correlations among semantically related sensor features through localized convolutions. Unlike feed-forward encoders that process each feature dimension independently, the CNN’s receptive fields encourage learning structured feature interactions within modality-specific groups. This inductive bias is especially beneficial under client heterogeneity, where preserving and leveraging local feature structure improves robustness to distributional differences across clients.

\subsubsection{Impact of Aggregation Strategies}
FedAvg~\cite{mcmahan2017communication}, FedProx~\cite{li2020federated_a}, and FedAdam~\cite{reddi2020adaptive} achieve largely comparable performance across all experiments (Table \ref{tab:avg_methods_results}). Among them, FedAvg attains a slightly higher AUROC, indicating that standard averaging is sufficient for stable convergence in our setting. Although FedProx is designed to mitigate client drift under high statistical heterogeneity or sporadic participation, its advantages are limited here because all six countries participate consistently in every round with relatively balanced contributions (Figure \ref{fig:3label}). Similarly, FedAdam employs adaptive server-side optimization that can accelerate early-stage convergence; however, it does not significantly improve final performance when clients already converge stably under FedAvg. These observations are consistent with prior findings by Khalil et al.~\cite{khalil2024exploring}, who reported that FedAvg remains competitive across diverse mental health FL applications due to its simplicity and robustness. Overall, our results indicate that model architecture and personalization strategy have a stronger impact on performance than the choice of aggregation method for cross-country mood inference.

\section{Limitations And Future Work}
While the proposed FedFAP framework emphasizes privacy-preserving learning by avoiding centralized data collection, it does not provide formal guarantees against information leakage from shared model updates. Recent work~\cite{zhu2019deep} has shown that intermediate representations and gradients can, in some cases, be exploited to infer sensitive attributes or reconstruct aspects of the original data. As such, privacy risks are reduced but not entirely eliminated. Future work should incorporate stronger privacy preserving mechanisms, such as secure aggregation~\cite{bonawitz2017practical} and differential privacy~\cite{geyer2017differentially}, to provide quantifiable privacy guarantees while maintaining predictive performance. Integrating these techniques would further strengthen the suitability of federated mood-aware systems for deployment in privacy-sensitive and regulated settings.

In this study, we assume a synchronous federated learning setup, where all selected clients participate in each communication round. However, real-world mobile and cross country deployments are often characterized by intermittent connectivity, resource constraints, and client dropouts. We do not explicitly evaluate the robustness of FedFAP under partial participation or asynchronous training scenarios. Exploring client availability dynamics, straggler mitigation, and asynchronous federated optimization remains an important direction for future work to better reflect real world deployment conditions.

Our evaluation primarily focuses on country-level analysis, which captures population-scale heterogeneity but may obscure finer-grained variability at the individual level. While country-wise analysis is useful for understanding cross-cultural and regional differences, mood inference is inherently personal, and individuals within the same country may exhibit substantially different behavioral patterns. Future work should extend this framework to participant-level personalization, particularly in settings where individual users contribute sufficient longitudinal data. Finally, deploying the proposed system in real-time, in-the-wild environments represents a critical next step. Such deployments would surface practical challenges related to energy consumption, latency, user consent, and long-term engagement, providing valuable insights that cannot be fully captured through offline evaluation alone.

\section{Conclusion}
Mood inference using mobile sensing is particularly challenging in cross-country settings, where behavioral patterns, sensing availability, and data distributions vary substantially across populations. To address these challenges, we proposed FedFAP, a feature-aware personalized federated learning framework that enables collaborative mood inference while preserving local data and accommodating feature heterogeneity. Our results demonstrate that FedFAP achieves strong performance, reaching an AUROC of \textbf{0.744} and outperforming both existing personalized federated baselines and centralized training setups evaluated in this study. These findings highlight the importance of personalization and feature-aware modeling for mood inference from mobile sensing data, and suggest that federated learning offers a practical and effective foundation for deploying privacy-aware, scalable mood-aware systems across diverse real-world contexts.

\bibliographystyle{ACM-Reference-Format}
\bibliography{sample-base}
\end{document}